\documentclass[letterpaper]{article}
\usepackage{aaai2027}
\usepackage[hyphens]{url}
\usepackage{graphicx}
\usepackage{natbib}
\usepackage{caption}
\usepackage{amsmath,amssymb,amsthm}
\usepackage{booktabs}
\usepackage{multirow}
\usepackage{xspace}
\newtheorem{theorem}{Theorem}
\newtheorem{proposition}[theorem]{Proposition}
\newcommand{\Prb}{\mathbb{P}}
\newcommand{\E}{\mathbb{E}}
\newcommand{\etal}{\emph{et al.}\xspace}
\newcommand{\cwnps}{cw-NPS\xspace}
\newcommand{\bnps}{b-NPS\xspace}

\title{NumBench: Diagnosing Counting Failures in Text-to-Image Models}

\author{
Sandeep Wadhwa$^{1}$,
Mayank Vatsa$^{1}$,
Richa Singh$^{1}$,
Parrva Chirag Shah$^{1}$,
Prakhar Galriya$^{2}$
}

\affiliations{
$^{1}$Indian Institute of Technology Jodhpur, India\\
$^{2}$Shiv Nadar University, India
}

\begin{document}
\maketitle

\begin{abstract}
Text-to-image (T2I) models often generate the wrong number of objects, yet existing benchmarks are too small or weakly controlled to explain why. We introduce \textbf{NumBench}, a benchmark of 640{,}000 prompts spanning 1{,}600 categories and counts from 1 to 100. Its factorial design varies object composition, spatial guidance, and appearance conditions while balancing counts and category exposure. We also develop a process model in which requested instances compete for a finite set of resolvable image regions. The model predicts a near-quadratic collision deficit at low occupancy and shows how coordinated placement reduces it. For scalable evaluation, we propose the Confidence-Weighted Numeric Precision Score (\cwnps), which aggregates three calibrated detectors and discounts uncertain proposals. Across five commercial systems, two open models, and two specialized counting methods, performance declines sharply with requested count; all evaluated methods are weak above 50 objects. Count range has the largest measured effect, followed by layout and composition. Grid guidance is strongest among guided layouts, consistent with the coordination prediction, although the analysis does not establish collision as the sole cause. A 14{,}400-image human study supports automated evaluation through count 50, while results on 243 natural-language prompts show transfer beyond NumBench templates.
\end{abstract}

\section{Introduction}
\label{sec:intro}

\begin{figure}[t]
\centering
\includegraphics[width=0.95\columnwidth]{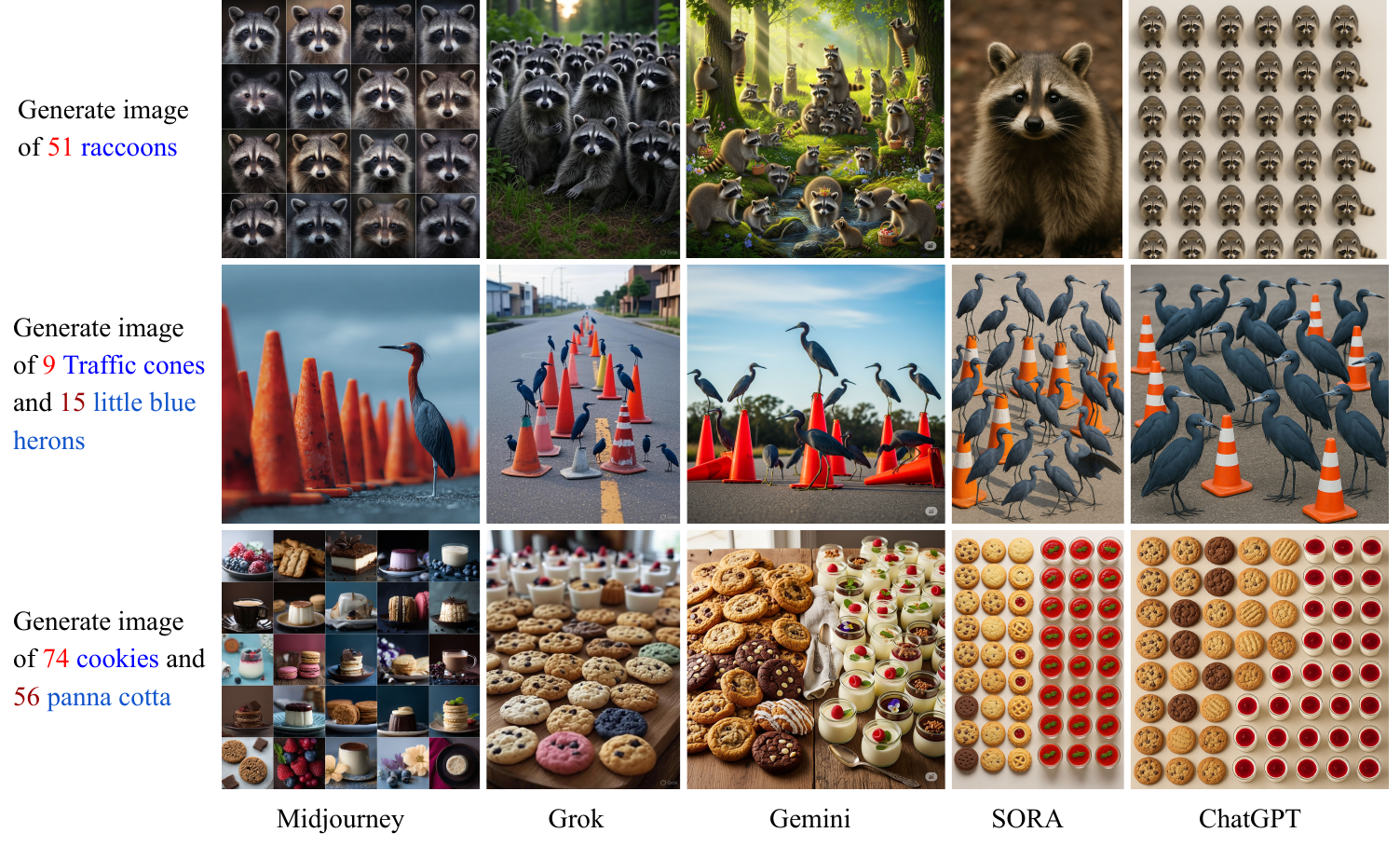}
\caption{Counting failures across five commercial T2I systems. High-count prompts produce missing or merged objects, category confusion, and repeated default patterns. System labels identify the evaluated interfaces; exact collection dates are recorded in the release metadata.}
\label{fig:teaser}
\end{figure}

Counting remains a persistent failure mode of modern T2I systems. A prompt such as ``generate 51 raccoons'' often yields far fewer objects; prompts requesting two categories additionally produce category mixing, merged instances, and default scene sizes (Figure~\ref{fig:teaser}). These failures matter wherever generated images must be numerically correct, including educational content, synthetic training data, accessibility tools, and digital commerce. Even when deployment counts are smaller, extending the range to 100 exposes where models cease to preserve distinct instance evidence and separates low-count success from scalable numeric control.

Existing evaluation does not isolate the causes. General compositional benchmarks include only small counting subsets~\cite{saharia2022imagen,ghosh2023geneval,huang2023t2icompbench,bakr2023hrsbench,hu2023tifa}; counting-specific sets remain limited in count range, category diversity, or controlled spatial variation~\cite{kajic2024geckonum,binyamin2025countgen,sun2024quota,zafar2024detection}. Moreover, detector-based metrics can confuse generator errors with detector misses, especially in crowded scenes.

We address this gap with three contributions. \textbf{First}, NumBench provides 640{,}000 prompts over 1{,}600 categories and counts 1--100. A controlled design crosses single- versus double-category composition, free-form versus guided layout, and clean versus modified appearance instructions. \textbf{Second}, we introduce a process model for spatial collision. It predicts how limited spatial capacity creates rapidly growing undercount and how coordinated layouts attenuate this loss. \textbf{Third}, we propose \cwnps, a calibrated detector-ensemble metric, and evaluate nine systems at substantially larger scale than prior counting studies.

The evaluation highlights a consistent hierarchy. Requested count is the dominant source of difficulty, followed by layout, composition, and appearance conditions. Specialized methods improve low-count results but remain weak at high counts. Among guided layouts, grids perform best, consistent with the coordination prediction of the process model. Human agreement and detector--human correlation weaken above count 50, so we treat that region as a stress test rather than a fully validated count estimate, and we release prompts, detector outputs, and annotations to support evaluation under the same protocol.

\begin{table*}[t]
\centering
\small
\setlength{\tabcolsep}{3.5pt}
\begin{tabular}{lccccccccc}
\toprule
\textbf{Benchmark} & \textbf{Prompts} & \shortstack{\textbf{Max}\\\textbf{Count}} & \textbf{Categories} & \textbf{Count} & \textbf{Layout} & \textbf{Appear.} & \textbf{Comp.} & \textbf{Joint} \\
\midrule
DrawBench$^\dagger$~\cite{saharia2022imagen} & $\sim$200 & $\sim$10 & $<20$ & $\times$ & $\times$ & $\times$ & $\times$ & $\times$ \\
PaintSkills$^\dagger$~\cite{cho2022dalleval} & 7{,}330 & $\sim$5 & $\sim$80 & $\times$ & $\times$ & $\times$ & $\times$ & $\times$ \\
GenEval$^\dagger$~\cite{ghosh2023geneval} & 553 & 4 & 80 & $\checkmark$ & $\times$ & $\times$ & $\times$ & $\times$ \\
T2I-CompBench$^\dagger$~\cite{huang2023t2icompbench} & 6{,}000 & $\sim$5 & $\sim$200 & $\times$ & $\times$ & $\times$ & $\times$ & $\times$ \\
HRS-Bench$^\dagger$~\cite{bakr2023hrsbench} & 45{,}000 & $\sim$10 & 700+ & $\times$ & $\times$ & $\times$ & $\times$ & $\times$ \\
GeckoNum~\cite{kajic2024geckonum} & 1{,}386 & $\sim$10 & $\sim$30 & $\checkmark$ & $\times$ & $\times$ & $\checkmark$ & $\times$ \\
Zafar \etal~\cite{zafar2024detection} & 3{,}674 & 25 & 147 & $\checkmark$ & $\times$ & $\times$ & $\times$ & $\times$ \\
CoCoCount~\cite{binyamin2025countgen} & $\sim$200 & $\sim$10 & $\sim$80 & $\checkmark$ & $\times$ & $\times$ & $\checkmark$ & $\times$ \\
QUANT-Bench~\cite{sun2024quota} & 475 & 25 & 19 & $\checkmark$ & $\times$ & $\times$ & $\times$ & $\times$ \\
\midrule
\textbf{NumBench} & \textbf{640{,}000} & \textbf{100} & \textbf{1{,}600} & $\checkmark$ & $\checkmark$ & $\checkmark$ & $\checkmark$ & $\checkmark$ \\
\bottomrule
\end{tabular}
\caption{Comparison with counting benchmarks. $^\dagger$Counting is one component of a broader suite. Joint indicates that count, layout, appearance, and composition are controlled within one benchmark design.}
\label{tab:benchmark}
\end{table*}

\section{Related Work}
\label{sec:related}

\noindent\textbf{Benchmarks.}
DrawBench~\cite{saharia2022imagen}, PaintSkills~\cite{cho2022dalleval}, GenEval~\cite{ghosh2023geneval}, T2I-CompBench~\cite{huang2023t2icompbench}, HRS-Bench~\cite{bakr2023hrsbench}, and TIFA~\cite{hu2023tifa} evaluate counting as one component of broader compositionality. Focused resources include GeckoNum~\cite{kajic2024geckonum}, CoCoCount~\cite{binyamin2025countgen}, QUANT-Bench, introduced with QUOTA~\cite{sun2024quota}, and the FSC-147-based protocol of Zafar \etal~\cite{zafar2024detection}. None jointly spans counts 1--100, 1{,}600 categories, composition, layout, and appearance variation at NumBench scale.

\noindent\textbf{Evaluation and generation.}
GenEval uses detection-based evaluation~\cite{ghosh2023geneval}, while TIFA uses visual question answering~\cite{hu2023tifa}. Most counting work relies on a single detector with a fixed threshold~\cite{binyamin2025countgen,zafar2024detection}. General-purpose evaluators such as GenEval and TIFA assess broader prompt--image consistency; our direct baseline is b-NPS because it represents the fixed-threshold detector-counting protocol used in prior exact-count evaluations. Generation methods modify count conditioning~\cite{binyamin2025countgen,kang2025counting}, use detector feedback~\cite{zafar2024detection}, plan scenes with language models~\cite{lian2024llmgrounded,wu2023selfcorrecting}, or impose spatial control~\cite{li2023gligen,zhang2023controlnet,dahary2024bounded}. NumBench supplies a common diagnostic testbed for these approaches.

\section{A Process Model for Spatial Collisions}
\label{sec:theory}

Consider a prompt requesting $n$ instances of one category. At the rendered object scale, suppose the image offers $M$ resolvable regions. Assigning two requested instances to the same region may merge them into one connected visible instance. Let $R$ be the number of occupied regions after placement.

\begin{theorem}[Collision deficit]
\label{thm:collision}
If $n\le M$ instances are assigned independently and uniformly to $M$ regions, then
\begin{equation}
\Prb(R=n)=\prod_{i=0}^{n-1}\!\left(1-\frac{i}{M}\right)
\le e^{-n(n-1)/(2M)},
\end{equation}
and
\begin{equation}
\E[n-R]=\frac{\binom{n}{2}}{M}+O\!\left(\frac{n^3}{M^2}\right).
\label{eq:collision}
\end{equation}
\end{theorem}

Thus, when $n/M$ is small, the expected visible deficit grows approximately quadratically rather than linearly. The model describes one mechanism only: it cannot explain overcounting, category substitution, or a generator's learned preference for common scene sizes.

To represent spatial coordination, let a placement use an unused region with probability $\lambda$ and otherwise choose uniformly, where $q=1-\lambda$.

\begin{proposition}[Coordination]
\label{prop:coordination}
For $q>0$,
\begin{align}
\E[R] &= \frac{M}{q}\!\left[1-\left(1-\frac{q}{M}\right)^n\right],\\
\E[n-R] &= q\frac{\binom{n}{2}}{M}+O\!\left(\frac{n^3}{M^2}\right).
\end{align}
For $\lambda=1$, every instance occupies an unused region and $R=n$.
\end{proposition}

Structured layout guidance should improve counting when it increases $\lambda$ or effective capacity $M$. This yields a testable qualitative prediction: regular layouts should degrade more slowly than weakly coordinated layouts. Our experiments test the predicted layout ordering through numeric precision and treat agreement as diagnostic consistency rather than causal identification.

\paragraph{Proof intuition.}
Let $I_u$ indicate whether region $u$ is occupied. A region remains empty with probability $(1-1/M)^n$, so linearity of expectation gives $\E[R]=M[1-(1-1/M)^n]$. Expanding the power yields the leading deficit $\binom{n}{2}/M$, which counts pairwise competition for the same region. With coordination, the recurrence for expected occupancy replaces $1/M$ by $q/M$; hence the leading deficit is multiplied by $q=1-\lambda$. Complete proofs are provided in the supplementary material.

The model yields two additional implications. First, increasing canvas resolution helps only when it increases the number of usable regions at the rendered object scale; upsampling after generation should not recover merged instances. Second, a layout method may improve counting by increasing $M$, increasing $\lambda$, or both. These mechanisms are observationally distinct only with directional count errors and controlled object scale. NumBench supplies the factors needed for such future tests, while the present paper uses the model to motivate and interpret the layout axis.

\section{NumBench}
\label{sec:dataset}

NumBench begins with 160{,}000 core prompts: 80{,}000 single-category and 80{,}000 double-category prompts. The vocabulary contains 1{,}600 unique categories curated from ImageNet, COCO, Objects365, Caltech-101, Caltech-256, Food101, and CUB-200. Synonyms, plural variants, and formatting duplicates are removed.

\begin{table}[t]
\centering
\small
\setlength{\tabcolsep}{3.2pt}
\begin{tabular}{lcl}
\toprule
\textbf{Dimension} & \textbf{Levels} & \textbf{Allocation} \\
\midrule
Composition & 2 & 80K single + 80K double \\
Count & 100 & Equal mass at each integer \\
Category & 1{,}600 & Marginally balanced assignment \\
Spatial layout & 4 & Free-form / grid / scene / random \\
Appearance & 2 & Clean + modified \\
\midrule
\textbf{Total} & & \textbf{640{,}000 prompts} \\
\bottomrule
\end{tabular}
\caption{NumBench construction. Every core prompt receives a free-form version, one guided counterpart, and clean and modified appearance forms.}
\label{tab:design}
\end{table}

Each single-category prompt pairs one category with a count in $[1,100]$. Every category appears 50 times and every integer appears 800 times. Each double-category prompt pairs two distinct categories with independently assigned counts; every category appears 100 times and every integer appears 1{,}600 times as a requested count. Category and count assignments are balanced marginally rather than forming a complete category–count crossing.

The core set is expanded along two axes. \textbf{Layout} adds a free-form version and a guided version using \texttt{grid}, \texttt{scene}, or \texttt{random}; guided forms are balanced across the three descriptors. After appearance expansion, free-form contributes 320{,}000 prompts and each guided form contributes approximately 106{,}667. Every core prompt therefore has a free-form reference but only one guided counterpart; factor summaries compare the four layout levels separately rather than pooling guided prompts. \textbf{Appearance} adds a clean and a modified version. Modified prompts combine two or three terms from \emph{blurry}, \emph{deformed}, \emph{low quality}, \emph{overlapping}, \emph{merged}, \emph{partial}, \emph{cropped}, and \emph{artifacts}. The final total is $160\mathrm{K}\times2\times2=640\mathrm{K}$ prompts. The scale comes from balanced coverage of count, category, composition, layout, and appearance cells rather than repeated paraphrases of one condition. NumBench is a controlled diagnostic distribution rather than an estimate of everyday prompting; the Experiments section includes a natural-language transfer test.

\noindent\textbf{Vocabulary and appearance semantics.}
Category names are converted to lowercase singular forms; clear synonyms, plural variants, and formatting duplicates are merged, while ambiguous names are removed when one phrase denotes unrelated concepts. Source labels and dataset provenance remain attached to every normalized category. The appearance axis contains two semantically different groups (Table~\ref{tab:appearance}). Separating them matters because one group mainly degrades image quality, whereas the other changes the visible evidence from which either a detector or a person must infer cardinality.

\begin{table}[t]
\centering
\small
\setlength{\tabcolsep}{3.4pt}
\renewcommand{\arraystretch}{1.06}
\begin{tabular}{p{1.55cm}p{2.45cm}p{3.05cm}}
\toprule
\textbf{Group} & \textbf{Descriptors} & \textbf{Interpretation} \\
\midrule
Visual quality & blurry, deformed, low quality, artifacts & Primarily changes fidelity and detector confidence. \\
Instance evidence & overlapping, merged, partial, cropped & Can make the requested cardinality visually ambiguous. \\
\bottomrule
\end{tabular}
\caption{The modified-appearance condition combines quality degradation with instance-separation difficulty; analyses keep these roles conceptually distinct.}
\label{tab:appearance}
\end{table}

\section{Confidence-Weighted Numeric Precision}
\label{sec:metric}

A fixed-threshold count from one detector is vulnerable to detector-specific bias, confidence drift, and recall loss in crowded scenes. We use OW-DETR~\cite{gupta2022owdetr}, Grounding DINO~\cite{liu2023grounding}, and OWL-ViT~\cite{minderer2022owlvit}. Each detector is queried through its native category interface, and class-aware non-maximum suppression is applied independently for every requested category before aggregation. The release preserves category mappings, raw proposals, calibrated confidences, and retained boxes, allowing each reported score to be audited. Additional vocabulary-coverage details are provided in the supplement.

For reference, \bnps uses hard counts $n_{di}^{(0.5)}$ from each detector at threshold .5:
\begin{equation}
\mathrm{b\mbox{-}NPS}=\max\!\left(0,1-\frac{1}{k}\sum_{i=1}^{k}\frac{|D^{-1}\sum_d n_{di}^{(0.5)}-n_i|}{n_i}\right).
\end{equation}
It removes single-detector dependence but still treats a .51 detection as fully present and a .49 detection as absent.

Detector $d$ returns raw confidence $\rho^{\rm raw}_{dij}$ for proposal $j$ of requested category $i$. We first apply temperature scaling~\cite{guo2017calibration},
\begin{equation}
\rho^{\rm sc}_{dij}=\sigma\!\left(\operatorname{logit}(\rho^{\rm raw}_{dij})/\tau_d\right),
\end{equation}
with temperatures fitted on a disjoint 5{,}000-image calibration set. The fitted values are 1.15, 1.42, and 0.93 for OW-DETR, Grounding DINO, and OWL-ViT. We then use the pre-specified reference-conditioned threshold
\begin{equation}
T(n)=\begin{cases}.7,&n\le3,\\ .3,&n\ge10,\\ .7-.4(n-3)/7,&\text{otherwise},\end{cases}
\end{equation}
and form the soft ensemble count $\hat n_i=D^{-1}\sum_d\sum_j \rho^{\rm sc}_{dij}\mathbf{1}[\rho^{\rm sc}_{dij}\ge T(n_i)]$. For $k$ requested categories,
\begin{equation}
\mathrm{cw\mbox{-}NPS}=\max\!\left(0,1-\frac{1}{k}\sum_{i=1}^{k}\frac{|\hat n_i-n_i|}{n_i}\right).
\label{eq:cwnps}
\end{equation}
The score lies in $[0,1]$. Calibration and thresholding are fixed before evaluating any of the nine systems. We pre-specify one threshold schedule and do not claim that it is optimal; paired b-NPS results test whether the main conclusions depend on confidence weighting alone. Because $T(n)$ uses the requested count, \cwnps is an evaluation score, not an independent count estimator. The schedule begins changing above count 3 and reaches .3 at count 10. It favors proposal recall in crowded scenes and can make signed errors less negative; a fixed .5 threshold has the opposite risk when crowding lowers detector confidence. Accordingly, the paper uses \cwnps for absolute numeric precision but does not infer a universal undercount direction from it.

\section{Experiments}
\label{sec:experiments}

\noindent\textbf{Systems and protocol.}
We evaluate five commercial systems (OpenAI image generation, Gemini, Sora, Midjourney, and Grok), two open models (SDXL and FLUX.2 [dev]), and two specialized methods (CountGen~\cite{binyamin2025countgen} and Bounded Attention~\cite{dahary2024bounded}). Commercial outputs were collected between January 2025 and July 2026 using default settings; per-system collection windows are recorded in the release metadata. OpenAI outputs were generated in January--February 2025 via ChatGPT, which at that time used DALL-E 3. Open and specialized systems use all 640{,}000 prompts with four images per prompt; commercial systems use a balanced 3{,}200-prompt subset with four images per prompt. All outputs are $1024\times1024$. The four locally run systems produce 10.24 million images; commercial systems add 64{,}000. We cache detector proposals so \bnps and \cwnps use identical raw outputs. The commercial subset is stratified by count range, composition, layout, and appearance. Factor effects are reported as the average best--worst difference across factor levels, averaged over the nine systems, rather than pooling all layout prompts into one guided category.

\begin{table*}[t]
\centering
\scriptsize
\setlength{\tabcolsep}{2.35pt}
\renewcommand{\arraystretch}{1.04}
\begin{tabular}{ll*{8}{cc}}
\toprule
& & \multicolumn{8}{c}{\textbf{Single Category}} & \multicolumn{8}{c}{\textbf{Double Category}} \\
\cmidrule(lr){3-10}\cmidrule(lr){11-18}
& & \multicolumn{2}{c}{1--25} & \multicolumn{2}{c}{26--50} & \multicolumn{2}{c}{51--75$^\ddagger$} & \multicolumn{2}{c}{76--100$^\ddagger$} &
\multicolumn{2}{c}{1--25} & \multicolumn{2}{c}{26--50} & \multicolumn{2}{c}{51--75$^\ddagger$} & \multicolumn{2}{c}{76--100$^\ddagger$} \\
\textbf{Class} & \textbf{Method} & b & cw & b & cw & b & cw & b & cw & b & cw & b & cw & b & cw & b & cw \\
\midrule
Open & SDXL & .602 & .553 & .402 & .360 & .250 & .222 & .099 & .076 & .483 & .445 & .301 & .271 & .152 & .125 & .051 & .038 \\
Open & FLUX.2 [dev] & .608 & .560 & .409 & .367 & .259 & .231 & .105 & .081 & .489 & .450 & .309 & .278 & .157 & .131 & .056 & .040 \\
\midrule
Specialized & CountGen & \textbf{.722} & \textbf{.692} & \textbf{.520} & \textbf{.485} & \textbf{.345} & \textbf{.311} & \textbf{.160} & \textbf{.129} & .642 & .610 & .443 & .410 & .255 & .230 & .111 & .088 \\
Specialized & Bounded Attention & .669 & .623 & .482 & .442 & .315 & .284 & .142 & .110 & \textbf{.671} & \textbf{.641} & \textbf{.465} & \textbf{.435} & \textbf{.279} & \textbf{.251} & \textbf{.125} & \textbf{.100} \\
\bottomrule
\end{tabular}
\caption{Basic (b) and confidence-weighted (cw) numeric precision on the full NumBench evaluation. Broad degradation and method ordering persist under both metrics. $^\ddagger$Values above count 50 are stress-test results because human agreement and detector--human correlation weaken in this range.}
\label{tab:main_results}
\end{table*}

\subsection{Performance Collapses as Counts Rise}

Every method declines across count ranges (Table~\ref{tab:main_results}). On single-category prompts, CountGen falls from .692 at counts 1--25 to .129 at 76--100; Bounded Attention falls from .623 to .110. Both open models fall below .09 in the final range. Specialized methods therefore help at low and moderate counts but do not solve high-count generation. CountGen loses 81.4\% of its single-category score between the first and last ranges; the corresponding loss for Bounded Attention is 82.3\%. The broad ordering is preserved under \bnps, indicating that the collapse is not created solely by confidence weighting. Across the table, \bnps is generally higher because hard thresholding treats every retained proposal as a full instance. On single-category prompts, the average b--cw gap across the four open and specialized methods is .043 for counts 1--25 and .028 for counts 76--100.

\begin{figure}[t]
\centering
\includegraphics[width=0.94\columnwidth]{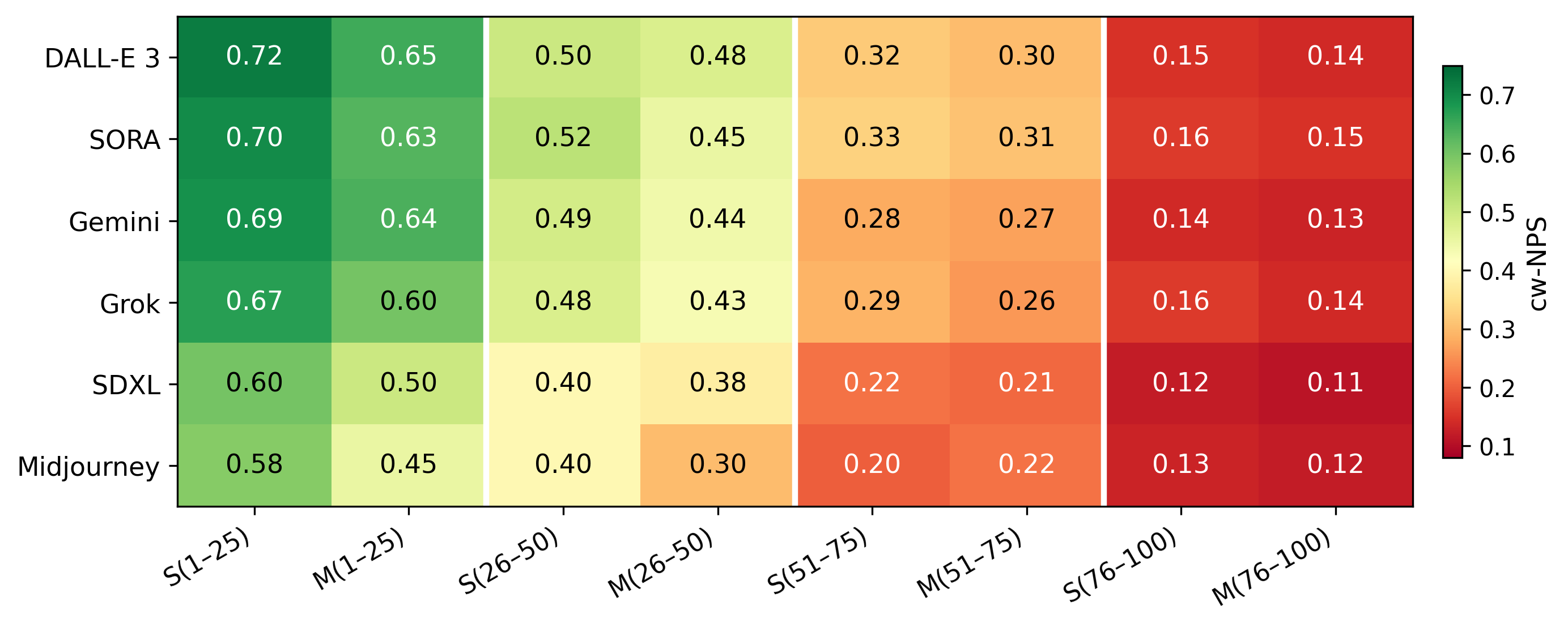}
\caption{cw-NPS on the balanced 3{,}200-prompt subset for five commercial systems, with SDXL included as an open-model reference. S and M denote single-category and two-category prompts, respectively.}
\label{fig:heatmap}
\end{figure}

The same trend appears for commercial systems (Figure~\ref{fig:heatmap}). The strongest displayed single-category score is below .75 in the 1–25 range, and all displayed systems are at or below .16 in the 76–100 range. Commercial results are snapshots because unversioned services can change without notice.

Composition changes method rankings. CountGen leads every single-category range, whereas Bounded Attention leads every double-category range. At counts 1--25, Bounded Attention reaches .641 on double-category prompts versus .610 for CountGen. Subject-wise spatial control may therefore help category separation, although double-category prompts are not uniformly harder for every system and range.

\subsection{Observed Sensitivity to Benchmark Factors}

\begin{figure*}[t]
\centering
\includegraphics[width=0.875\textwidth]{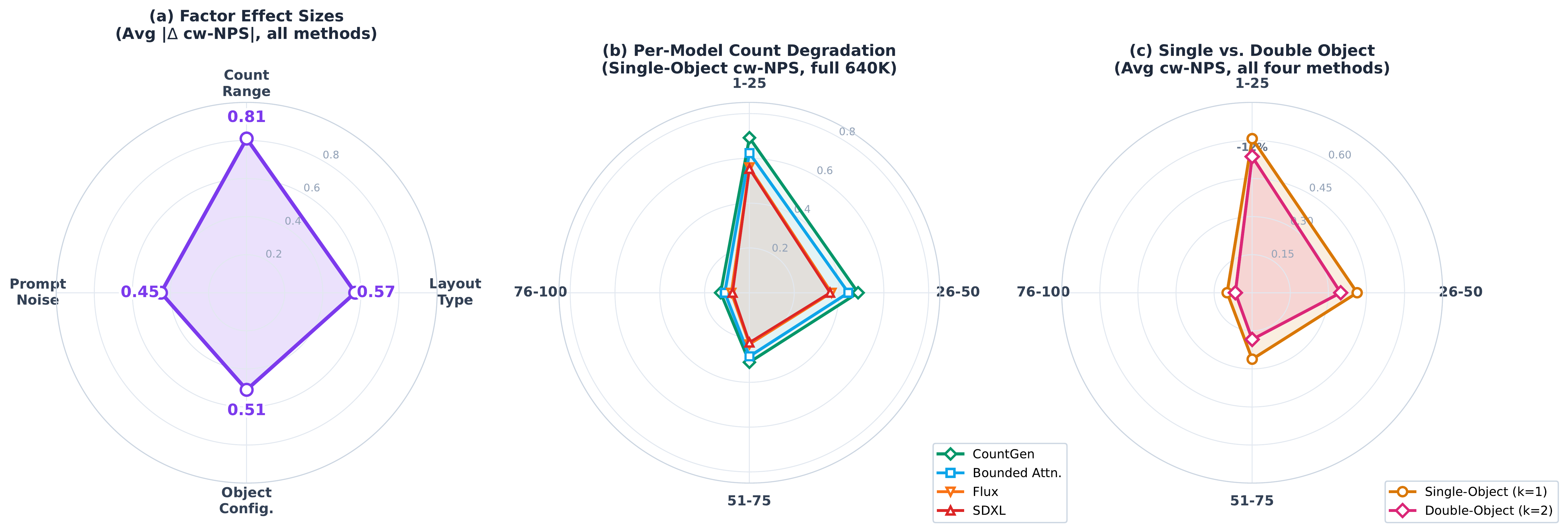}
\caption{Factor analysis. Axis and legend labels use shorthand: ``Prompt Noise'' denotes the appearance condition, ``Object Config.'' denotes composition, and ``Single/Double-Object'' denotes single- and double-category prompts. (a) Count range has the largest effect (normalized values). (b) Specialized methods improve low-count performance but not the high-count regime. (c) Composition changes difficulty across count ranges.}
\label{fig:radar}
\end{figure*}

We use the balanced design to measure descriptive score variation across factor levels. For factor $f$, within each cell defined by the remaining factors we compute the relative contrast between the best and worst levels of $f$, $(\mathrm{best}-\mathrm{worst})/\mathrm{best}$, and average these contrasts over cells and systems. The resulting values lie in $[0,1]$ and measure proportional sensitivity, not absolute cw-NPS differences. These values identify strong associations within NumBench; they are not causal coefficients. The resulting hierarchy is count range (.81), layout (.57), composition (.51), and appearance condition (.45). Figure~\ref{fig:radar} shows that specialized methods retain a low-count advantage but all systems collapse toward the origin above 50 objects.

Among the three guided settings, grid has the highest average \cwnps (.45), while scene (.31) and random (.32) are similar. This ordering is consistent with Proposition~\ref{prop:coordination}: regular placement can reduce competition for resolvable regions. The comparison remains observational and does not include the matched free-form aggregate needed for a complete four-layout contrast.

Under the cell-balanced aggregation defined above, appearance condition has a mean best--worst difference of .45. This aggregate is not a uniform .45 reduction within every count range. Blur, deformation, low quality, and artifacts primarily change visual quality; overlap, merging, cropping, and partial objects directly change the visible evidence for counting. The supplementary material separates these groups.

\subsection{Metric Validation and Transfer}
\label{sec:validation}

Replacing the ensemble with any single detector shifts aggregate \cwnps by at least .10, showing that detector choice materially affects measurement. The human study below evaluates the ensemble score.

Three annotators independently counted 1{,}600 images per system, yielding 14{,}400 images. Each count range contributes 400 images per system, with coverage across layouts, appearance conditions, and both composition settings. Table~\ref{tab:human_validation} summarizes the reported validation strata. Agreement is strongest through count 50, making this the better-validated evaluation regime. Results above 50 remain informative as stress tests, but their absolute values should be interpreted cautiously. Annotators saw the image and target category, counted distinct visible instances, and could mark ambiguous cases. Merged objects were counted by visible bodies when separable; cropped objects were counted only when a distinct instance remained clear. Consensus counts replace detector counts in Equation~\ref{eq:cwnps} to form the human score.

\begin{table}[t]
\centering
\small
\setlength{\tabcolsep}{4pt}
\renewcommand{\arraystretch}{1.06}
\begin{tabular}{lcc}
\toprule
\textbf{Count range} & Fleiss' $\boldsymbol{\kappa}$ & Detector--human $r$ \\
\midrule
1--25 & .82 & $>.80$ \\
26--50 & .58 & $>.80$ \\
76--100$^\ddagger$ & .33 & $\approx .40$ \\
Overall & --- & $>.75$ \\
\bottomrule
\end{tabular}
\caption{Aggregate human validation for the reported strata. The source annotations do not provide separate $\kappa$ or detector--human $r$ estimates for 51--75. $^\ddagger$High-count stress test.}
\label{tab:human_validation}
\end{table}

\begin{figure*}[t]
\centering
\includegraphics[width=0.975\textwidth]{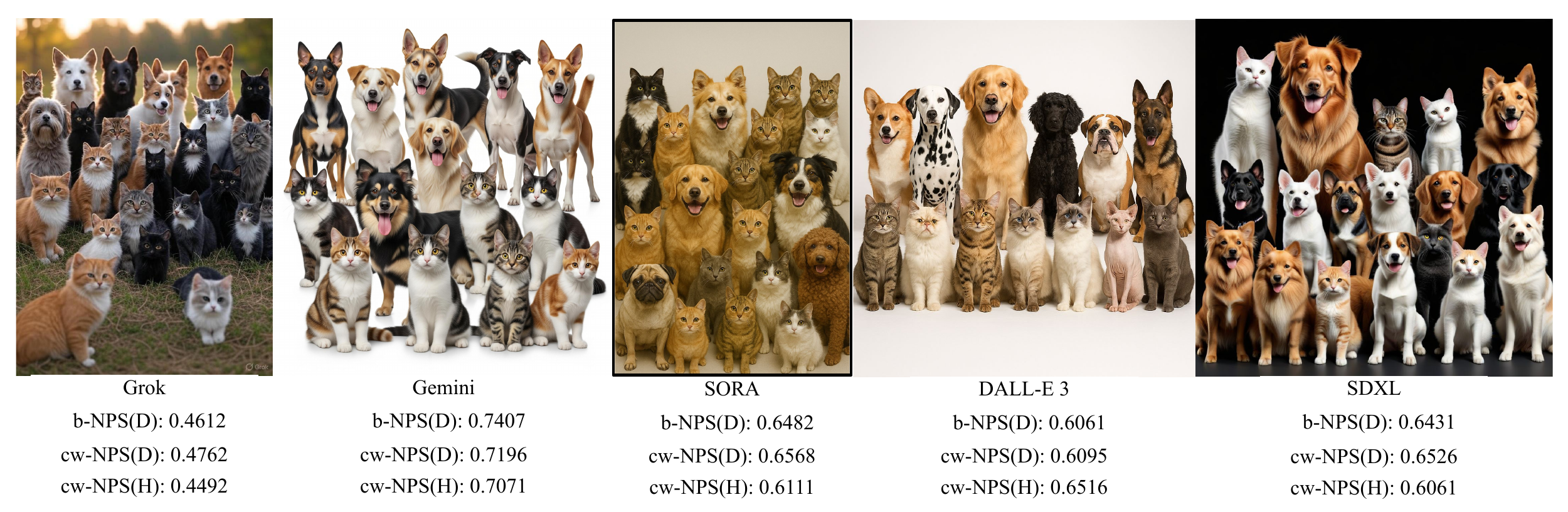}
\caption{Human validation for ``11 cats and 9 dogs.'' Detector-based and human-count \cwnps differ by at most .05 for each shown system. This example illustrates agreement in the validated count range; the aggregate correlation analysis, not a single image, is the main evidence.}
\label{fig:human}
\end{figure*}

Finally, we construct CoCoCount+ by combining 200 prompts from CoCoCount~\cite{binyamin2025countgen}, 19 counting prompts from DrawBench~\cite{saharia2022imagen}, and 24 prompts used in prior comparative studies~\cite{binyamin2025countgen,dahary2024bounded}. Its 243 natural-language prompts all request fewer than ten objects. Method rankings and performance gaps remain broadly consistent with NumBench, and no system exceeds .67 \cwnps. Qualitative cases also separate total-count errors from category-binding errors: a model may generate five kittens while assigning the wrong color ratio. The examples show that visually plausible outputs can still contain cardinality or category-binding errors, supporting a dedicated count metric rather than visual quality alone.

\noindent\textbf{Interpreting the high-count regime.}
Three uncertainties accumulate as requested count rises. First, generators increasingly merge, crop, or occlude instances, so even the intended visual unit becomes ambiguous. Second, detector recall and confidence decline as objects become smaller and more crowded. Third, human annotators disagree more often in the same scenes. The paired b-NPS/cw-NPS table shows that the broad performance collapse survives two counting rules, but neither rule converts visually ambiguous scenes into exact ground truth. We therefore use counts 51--100 to test whether a method preserves usable numeric evidence under severe crowding, not to claim equally precise measurement across all 100 counts. This distinction also explains why the process model is interpreted through layout ordering and absolute precision rather than through an unqualified signed-undercount claim.

\begin{figure*}[]
\centering
\includegraphics[width=0.9\textwidth]{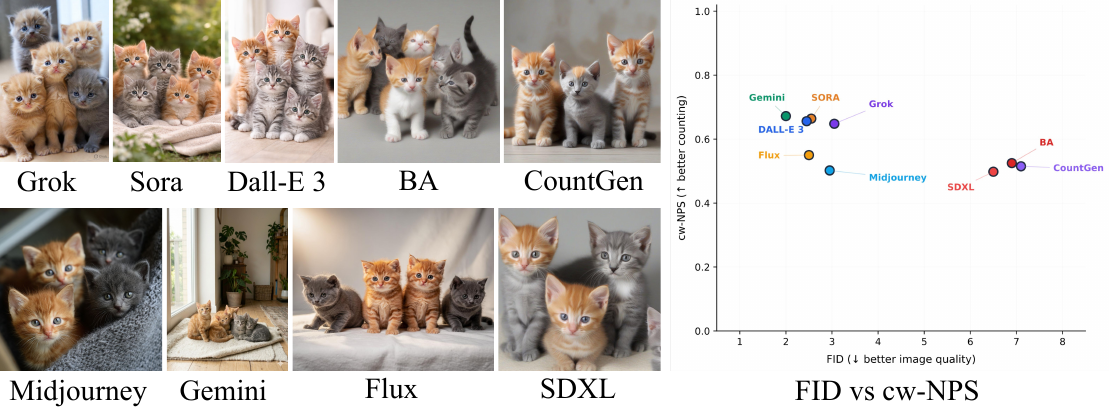}
\caption{Transfer to CoCoCount+. Left: generations for ``3 ginger kittens and 2 grey kittens'' expose exact-count, color-binding, missing-object, and overcount failures even in the low-count regime. Right: FID versus cw-NPS across the nine systems; perceptual quality does not predict numeric precision, and no evaluated system exceeds .67 cw-NPS on the 243 prompts.}
\label{fig:external}
\end{figure*}

\subsection{What the Benchmark Diagnoses}

NumBench separates three failure families that are often collapsed into one score. \textbf{Cardinality errors} change the number of visible target instances. \textbf{Binding errors} preserve the total but assign instances to the wrong category or attribute, as in the color-ratio example in Figure~\ref{fig:external}. \textbf{Visibility errors} create partial, merged, or ambiguous instances that make both human and automated counting uncertain. Equation~\ref{eq:cwnps} evaluates category-wise cardinality and therefore penalizes many binding errors, but it cannot fully resolve ambiguous visibility without a stronger instance-level reference.

Category-wise scoring is essential. Suppose a prompt requests three cats and two dogs, but the image contains five cats and no dogs. A total-only evaluator reports an exact count of five. In contrast, the category-wise score is $1-\tfrac{1}{2}(2/3+1)=1/6$ before clipping, correctly exposing the allocation failure. The same principle applies to attributes such as color when they define separate requested groups.

This distinction matters for model development. A count-conditioning method should be judged on cardinality after controlling layout and appearance. A spatial-control method should additionally report category binding on double-category prompts. An evaluator should distinguish its better-validated regime from high-count stress tests rather than presenting all scores as equally certain. NumBench supports all three analyses through factor labels, category-wise detector outputs, and human annotations.

\begin{table*}[t]
\centering
\small
\setlength{\tabcolsep}{4.2pt}
\renewcommand{\arraystretch}{1.08}
\begin{tabular}{p{1.55cm}p{3.2cm}p{3.0cm}p{3.45cm}p{3.15cm}}
\toprule
\textbf{Failure} & \textbf{Observable symptom} & \textbf{Most relevant axes} & \textbf{What NumBench measures} & \textbf{Recommended reporting} \\
\midrule
Cardinality & Too few or too many visible target instances & Count, layout & Category-wise absolute count error through b-NPS and cw-NPS & Scores by count range and composition \\
Binding & Correct total but wrong category or attribute allocation & Composition, category & Separate target counts for every requested category & Per-category counts on double-category prompts \\
Visibility & Merged, partial, cropped, or ambiguous instances & Appearance, layout & Detector confidence plus human ambiguity and agreement & Human validation and explicit stress-test labels \\
\bottomrule
\end{tabular}
\caption{NumBench separates three failure families that require different evidence and reporting. A single total-count score cannot distinguish them.}
\label{tab:diagnostic_use}
\end{table*}

Table~\ref{tab:diagnostic_use} clarifies how the benchmark should guide interventions. Prompt or count-conditioning methods primarily target cardinality; subject-wise layout controls should be examined for both cardinality and binding; and evaluator improvements should focus on visibility ambiguity and detector robustness. This prevents an apparent gain in one failure family from being presented as a complete solution to counting.

\noindent\textbf{Reproducibility and Release:} The complete 640{,}000-prompt dataset is uploaded as the supplementary submission. After acceptance, the full release will include every prompt and factor label, source codes, and meta-data associated with our experiments. 

\section{Limitations and Broader Impact}
\noindent\textbf{Measurement validity.}
\cwnps remains detector-mediated: its three detectors reduce single-model dependence but may share training-data biases, miss the same small objects, or confuse similar categories. The count-conditioned threshold favors recall in crowded scenes and can make signed errors less negative, whereas a fixed threshold can exaggerate undercount as confidence falls with object size; we therefore report absolute precision, pair it with b-NPS, and avoid claiming a universal error direction. The human study provides an independent reference, but human counting also destabilizes in dense scenes, so results above 50 should be read as stress tests of preserved numeric evidence rather than equally precise count estimates.

\noindent\textbf{Model and construct validity.}
The collision process is a deliberately simple null model: it assumes a fixed effective capacity $M$ and explains only merging-induced deficits, not overcounting, category substitution, object-scale or scene-structure changes, or learned scene-size priors. Grid guidance is observationally consistent with stronger coordination, but the present comparison cannot separate larger effective capacity, less overlap, larger objects, simpler backgrounds, or improved detector confidence. The model is thus a testable mechanism proposal, not a complete account of counting failure.
\noindent\textbf{Dataset and temporal validity.}
Controlled templates support paired factor analysis but exclude paraphrases, number words, multilingual instructions, and long natural-language prompts; CoCoCount+ offers an external check but contains only 243 low-count prompts. The appearance axis intentionally mixes quality degradation with instance-evidence modification, so pooled effects should not be read as one homogeneous corruption, and categories inherit the visual scope of their source datasets. Commercial systems are unversioned snapshots collected between January 2025 and July 2026, and later service updates may change their results. These boundaries motivate releasing prompts, timestamps, detector proposals, and open-model outputs so future systems can be evaluated under the same protocol.

\paragraph{Ethical Statement:}
NumBench is intended to improve evaluation for applications requiring visual correctness. The work introduces no new image-generation capability. Categories are drawn from established recognition datasets, with offensive labels and sensitive personal attributes excluded. Commercial outputs are redistributed only when platform terms permit. The complete prompt dataset is included with the supplementary submission. Metadata, open-model outputs, detector proposals, calibration code, and evaluation scripts will be released under a permissive license upon acceptance.

\section{Conclusion}
NumBench turns T2I counting into a controlled diagnostic problem spanning counts 1--100, 1{,}600 categories, and 640{,}000 prompts. A spatial-collision model explains why limited coordination can produce rapidly growing visible deficits, while \cwnps supports evaluation at scale. Across nine systems, requested count dominates all measured factors; layout and composition also matter, and no evaluated method retains strong numeric precision at high counts. Beyond aggregate scores, the benchmark separates cardinality, binding, and visibility failures, so a method can be credited for the specific failure family it addresses rather than for counting as a whole. Human and external evaluations support scalable use in the low-to-mid count regime, while higher counts remain stress tests, and the released prompts, detector proposals, and annotations allow every reported score to be audited and future systems to be evaluated under the same protocol. Progress requires coordinating many distinct instances while preserving category binding and countable visual evidence.

\label{page:lastcontent}
\bibliography{numbench}

\end{document}


\maketitle

This supplement provides complete proofs, benchmark-construction and metric-implementation details, the human-evaluation protocol, natural-language transfer details, and additional limitations and release information. The paired b-NPS/cw-NPS results and the aggregate human-validation table are reported in the main paper.

\section{Proofs for the Spatial-Collision Model}

\subsection{Collision Deficit}
The main paper considers $n$ requested instances assigned independently and uniformly to $M$ resolvable regions, and lets $R$ denote the number of occupied regions. All assignments are distinct only when each new instance avoids the regions already occupied. Therefore,
\begin{equation}
\Prb(R=n)=\prod_{i=0}^{n-1}\left(1-\frac{i}{M}\right).
\end{equation}
Using $1-x\leq e^{-x}$ gives
\begin{align}
\Prb(R=n)
&\leq \exp\left(-\frac{1}{M}\sum_{i=0}^{n-1}i\right)\\
&=\exp\left(-\frac{n(n-1)}{2M}\right).
\end{align}

Let $I_u$ indicate whether region $u$ is occupied. A fixed region is empty with probability $(1-1/M)^n$. By linearity of expectation,
\begin{equation}
\E[R]=\sum_{u=1}^{M}\E[I_u]
=M\left[1-\left(1-\frac{1}{M}\right)^n\right].
\end{equation}
The binomial expansion then gives
\begin{align}
\E[n-R]
&=\frac{\binom{n}{2}}{M}-\frac{\binom{n}{3}}{M^2}
+O\left(\frac{n^4}{M^3}\right)\\
&=\frac{\binom{n}{2}}{M}+O\left(\frac{n^3}{M^2}\right),
\end{align}
where the final form describes the leading low-occupancy behavior. Thus, collision-induced visible deficit grows approximately quadratically in $n$ when $n/M$ is small.

\subsection{Coordinated Placement}
Let a placement choose an unused region with probability $\lambda$ and otherwise choose uniformly, and write $q=1-\lambda$. Let $R_t$ be the number of occupied regions after $t$ placements. A coordinated step always increases occupancy by one. A uniform step succeeds with probability $1-R_{t-1}/M$. Hence,
\begin{equation}
\E[R_t\mid R_{t-1}]=R_{t-1}+1-\frac{q}{M}R_{t-1}.
\end{equation}
Writing $m_t=\E[R_t]$ yields the recurrence
\begin{equation}
m_t=\left(1-\frac{q}{M}\right)m_{t-1}+1,\qquad m_0=0.
\end{equation}
For $q>0$, its solution is
\begin{equation}
m_n=\frac{M}{q}\left[1-\left(1-\frac{q}{M}\right)^n\right].
\end{equation}
Expanding the power gives
\begin{equation}
\E[n-R]=q\frac{\binom{n}{2}}{M}
-q^2\frac{\binom{n}{3}}{M^2}
+O\left(\frac{n^4}{M^3}\right).
\end{equation}
Coordination therefore scales the leading collision deficit by $q=1-\lambda$. When $\lambda=1$, every placement uses a new region and $R=n$.

\paragraph{Scope of the model.}
The derivation describes one plausible spatial mechanism. It does not model overcounting, category substitution, object-scale changes, detector misses, or learned scene-size priors. The layout results in the main paper are consequently interpreted as diagnostic consistency, not causal proof.

\section{Benchmark Construction}

NumBench starts with 160,000 core prompts: 80,000 single-category prompts and 80,000 double-category prompts. Every core prompt is instantiated once without spatial guidance and once with one guided layout selected from grid, scene, and random placement. Each of these 320,000 layout prompts is then rendered in a clean and a modified appearance form, producing 640,000 prompts.

\begin{table}[t]
\centering
\footnotesize
\setlength{\tabcolsep}{3.5pt}
\renewcommand{\arraystretch}{1.08}
\begin{tabular}{@{}p{1.68cm}cp{1.62cm}p{2.95cm}@{}}
\toprule
\textbf{Factor} & \textbf{Levels} & \textbf{Allocation} & \textbf{Purpose} \\
\midrule
Count & 100 & Equal by integer & Resolve degradation without broad-bin confounding. \\
Layout & 4 & Free plus one guided form & Probe spatial organization. \\
Appearance & 2 & Clean and modified & Probe visual and instance-separation difficulty. \\
Composition & 2 & One or two categories & Probe count--category binding. \\
Category & 1,600 & Incomplete blocks & Broaden object coverage and category difficulty. \\
\midrule
\textbf{Total} & & \textbf{640,000} & \\
\bottomrule
\end{tabular}
\caption{NumBench factors. Counts and composition are balanced; categories are not completely crossed with every condition.}
\label{tab:supp_design}
\end{table}

\subsection{Count and Layout Allocation}
Every integer from 1 to 100 has equal mass within each composition setting. After appearance expansion, the free-form condition contains 320,000 prompts. The three guided layouts contain approximately 106,667 prompts each. This asymmetry is deliberate: every core prompt receives a free-form reference, but only one guided counterpart. Layout summaries therefore compare levels separately rather than treating all guided prompts as one pooled condition.

\subsection{Vocabulary Provenance and Appearance Metadata}
The vocabulary is drawn from ImageNet, COCO, Objects365, Caltech-101, Caltech-256, Food101, and CUB-200. The main paper summarizes normalization and the two appearance groups. The release retains the original source label, normalized category name, source dataset, selected appearance terms, and whether each term primarily concerns visual quality or visible instance evidence. These fields permit category-level and appearance-group analyses without reconstructing curation decisions.

\section{Metric Implementation}

\subsection{Basic Numeric Precision}
For $k$ requested categories with target counts $n_i$ and predicted counts $\hat n_i$, basic numeric precision is
\begin{equation}
\mathrm{b\mbox{-}NPS}=\max\left(0,1-\frac{1}{k}\sum_{i=1}^{k}\frac{|\hat n_i-n_i|}{n_i}\right).
\end{equation}
It returns one for exact counting and clips to zero once average relative error reaches one.

\subsection{Confidence-Weighted Numeric Precision}
The evaluator combines OW-DETR~\cite{gupta2022owdetr}, Grounding DINO~\cite{liu2023grounding}, and OWL-ViT~\cite{minderer2022owlvit}. Detector-specific category interfaces are used, and class-aware non-maximum suppression is applied independently for each requested category. The release will preserve the category mapping, raw proposals, calibrated confidences, and retained boxes for auditability.

For detector $d$ with logit $z_{d,j}$, calibrated confidence is
\begin{equation}
\tilde p_{d,j}=\sigma\left(\frac{z_{d,j}}{\theta_d}\right),
\end{equation}
where $\theta_d$ is fitted on a held-out 5,000-image calibration split. The fitted temperatures used in the evaluation are 1.15 for OW-DETR, 1.42 for Grounding DINO, and 0.93 for OWL-ViT. The reference-conditioned threshold is
\begin{equation}
T(n)=
\begin{cases}
0.7, & n\leq 3,\\
0.3, & n\geq 10,\\
0.7-0.4(n-3)/7, & \text{otherwise}.
\end{cases}
\end{equation}
After thresholding, the soft count averages summed confidence across the three detectors. Substituting these soft counts into the b-NPS equation gives \cwnps.

\paragraph{Interpretation.}
Because $T(n)$ uses the requested count, \cwnps is an evaluation score rather than a stand-alone counting algorithm. The lower threshold at larger $n$ improves proposal recall in crowded scenes, but it can also make signed error less negative. A fixed threshold can bias in the opposite direction when object confidence falls with crowding. The main paper therefore uses absolute error and does not infer a universal undercount direction from \cwnps alone.

\section{Additional Result Interpretation}

The main paper reports paired b-NPS and cw-NPS values for every open and specialized system, composition setting, and count range. Across those cells, cw-NPS is usually lower because it discounts uncertain proposals rather than counting every retained box as one complete instance. The average b--cw difference is .043 for counts 1--25 and .028 for counts 76--100. CountGen is strongest in all single-category ranges, whereas Bounded Attention is strongest in all double-category ranges. This reversal suggests that subject-wise spatial controls can improve category binding even when they do not solve high-count generation.

\section{Human-Evaluation Protocol}

Three annotators independently count 1,600 images from each of the nine systems, for 14,400 evaluated images. Each count range contributes 400 images per system, with coverage across layouts, appearance conditions, and both composition settings. Annotators see the image and target category, count distinct visible instances, and may mark ambiguous cases. Merged objects are counted by visible bodies when separable. Cropped objects count only when a distinct instance remains clear. Consensus counts replace detector counts in the cw-NPS equation to form the human score.

The aggregate agreement and detector--human correlation values are reported in the main paper. Agreement is strongest through count 50. At higher counts, dense overlap makes the reference count itself less stable, so the paper presents those results as stress tests rather than as equally validated absolute measurements. No separate Fleiss' $\kappa$ estimate is available for the 51--75 stratum; the main table therefore reports only the measured strata rather than interpolating the missing value.

\section{External Natural-Language Evaluation}

We also apply \cwnps to 243 natural-language prompts from CoCoCount+. All prompts request fewer than ten objects. Broad method rankings and performance gaps remain consistent with NumBench, and no evaluated system exceeds .67 \cwnps. This experiment tests transfer beyond the benchmark's controlled templates, although its small size and low count range do not substitute for the full factorial evaluation.

\section{Release Schema and Re-evaluation}

The main paper now gives the complete validity discussion. The release is structured to support re-evaluation rather than only reproduce aggregate tables. Each record contains the prompt, factor levels, source and normalized category labels, system name, collection timestamp, resolution, retry count, detector query, raw proposal, calibrated confidence, retained box, and final metric value. Local-model records additionally include exact checkpoint revisions, software versions, random seeds, sampling settings, and hardware. Calibration annotations remain separate from benchmark evaluation. Commercial outputs are redistributed only when platform terms permit; prompts and evaluation metadata remain available for later service versions.

\bibliography{numbench}